\documentclass[conference]{IEEEtran}
\usepackage{cite}
\usepackage{amsmath,amssymb,amsfonts}
\usepackage{algorithmic}
\usepackage{graphicx}
\usepackage{textcomp}
\usepackage{xcolor}
\usepackage{hyperref}       
\usepackage{url}            
\usepackage{booktabs}       
\usepackage{amsfonts}       
\usepackage{nicefrac}       
\usepackage{microtype}      
\usepackage{xcolor}         
\usepackage{graphicx}
\usepackage{pifont}
\usepackage{wrapfig}
\usepackage{multirow}
\usepackage{float}
\usepackage{adjustbox}

\newcommand{\name}{VeriTrace}

\newcommand{\cmark}{\ding{51}}
\newcommand{\xmark}{\ding{55}}

\def\BibTeX{{\rm B\kern-.05em{\sc i\kern-.025em b}\kern-.08em
    T\kern-.1667em\lower.7ex\hbox{E}\kern-.125emX}}
\begin{document}

\IEEEoverridecommandlockouts
\IEEEpubid{\makebox[\columnwidth]{ 979-8-3195-1246-8-0/26/\$31.00 \copyright2026 IEEE \hfill} \hspace{\columnsep}\makebox[\columnwidth]{ }}

\title{\name: Human-Like Temporal Exploration Completes Agentic Action Space}

\author{\IEEEauthorblockN{Yu-Tung Liu}
\IEEEauthorblockA{University of Maryland, College Park\\
College Park, USA \\
liuyt@umd.edu}
\and
\IEEEauthorblockN{Cunxi Yu}
\IEEEauthorblockA{University of Maryland, College Park\\
College Park, USA \\
cunxiyu@umd.edu}
}

\maketitle

\begin{abstract}
Large language models have shown promise for automated Verilog RTL generation, yet state-of-the-art multi-agent systems plateau at ~95\% accuracy on standard benchmarks. We trace this ceiling to an incomplete \emph{debugging action space}: existing systems restrict which signals the agent can inspect, which time windows it can query, or both, reducing debugging to pattern matching on a narrow, predetermined view of circuit behavior rather than hypothesis-driven root-cause analysis. We present \name, a multi-agent system whose Inspector agent operates over a \emph{complete} debugging action space, with independent control over signal selection, time-window bounds, and iteration depth. This capability, which we term \textbf{Agentic Temporal Exploration}, enables the agent to form hypotheses about failure causes, query the waveform for evidence, and refine its understanding iteratively, mirroring the exploratory process of human verification engineers. \name\ achieves \textbf{100\% Pass@1} on VerilogEval-V2, the first system to attain perfect functional correctness on this benchmark.
On a shared Claude Sonnet 4.0 backbone, VeriTrace outperforms the strongest reproduced baseline by +5.1\%, demonstrating that debugging agency closes the final accuracy gap.

\end{abstract}


\section{Introduction}

Modern digital hardware design remains a labor-intensive process that requires significant expertise in hardware description languages (HDLs) such as Verilog or SystemVerilog~\cite{choi2013hls, faber2022challenges}. As chip complexity grows and design cycles shorten, the demand for automated RTL code generation has become increasingly urgent~\cite{liu2023verilogeval, pan2025survey, lu2024rtllm}. Manual translation from natural language specifications to functionally correct Verilog is error-prone and time-consuming, motivating the need for intelligent automation.

Large language models (LLMs) have emerged as promising tools for hardware design automation~\cite{pan2025survey, thakur2022benchmarkinglargelanguagemodels}. Recent work has demonstrated that LLMs can generate syntactically valid Verilog code from natural language descriptions, with models like GPT-4 and Claude achieving non-trivial pass rates on benchmarks such as VerilogEval~\cite{liu2023verilogeval, pinckney2025revisiting}. These results suggest that LLMs possess sufficient understanding of HDL syntax and basic digital design patterns to serve as the foundation for automated RTL generation~\cite{pinckney2025revisiting, zehua2024betterv, tsai2024rtlfixerautomaticallyfixingrtl}.

However, single-agent LLM approaches face fundamental limitations in debugging. When generated code fails simulation, these systems receive only pass/fail feedback or generic error messages, lacking the detailed signal-level information needed to diagnose functional bugs. Multi-agent systems have begun to address this gap by incorporating waveform-based debugging feedback~\cite{ho2025verilogcoder, zhao2025mage}, but the feedback alone is not sufficient. What matters is the \emph{action space} available to the debugging agent: which signals it can select, which time windows it can query, and whether it can issue queries iteratively as its understanding evolves\cite{wu2024autogen, chen2023teaching}.

Table~\ref{tab: comp_method} summarizes the debugging action spaces of existing multi-agent systems. MAGE~\cite{zhao2025mage} introduces state checkpoint printing that captures I/O values at the first error timestamp, but restricts the agent to input and output signals at a single point in time, no signal selection, no temporal freedom, and no iteration.  VerilogCoder~\cite{ho2025verilogcoder} uses AST analysis to automate signal selection and supports iterative tracing, but anchors its waveform display to the first mismatch timestamp. Although both systems demonstrate that waveform feedback improves functional correctness, neither grants the agent joint control over \emph{which} signals to inspect \emph{and when}.

\begin{table}[h!]
    \centering
    \caption{Comparison of multi-agent systems.}
    \label{tab: comp_method}
\begin{tabular}{@{}cccc@{}}
\toprule
\multirow{2}{*}{Method} &
  \multirow{2}{*}{\begin{tabular}[c]{@{}c@{}}Agentic Signal\\ Selection\end{tabular}} &
  \multirow{2}{*}{\begin{tabular}[c]{@{}c@{}}Agentic Temporal\\ Exploration\end{tabular}} &
  \multirow{2}{*}{Iterative} \\
                 &               &   &   \\ \midrule
MAGE             & \xmark             & \xmark & \xmark \\
VerilogCoder     & \cmark (AST-based) & \xmark & \cmark \\
VeriTrace (Ours) & \cmark             & \cmark & \cmark \\ \bottomrule
\end{tabular}
\end{table}

We present \name, a multi-agent system that closes this action space gap by introducing \textbf{Agentic Temporal Exploration}, a debugging methodology that grants the LLM agent full control over signal selection and time-window inspection. Unlike prior approaches, \name\ enables iterative waveform queries: the agent selects signals and a time range, observes the result, refines its hypothesis, and queries again until the root cause is identified. Our contributions are as follows:

\begin{figure*}[!t]
  \centering
  \includegraphics[width=0.95\textwidth]{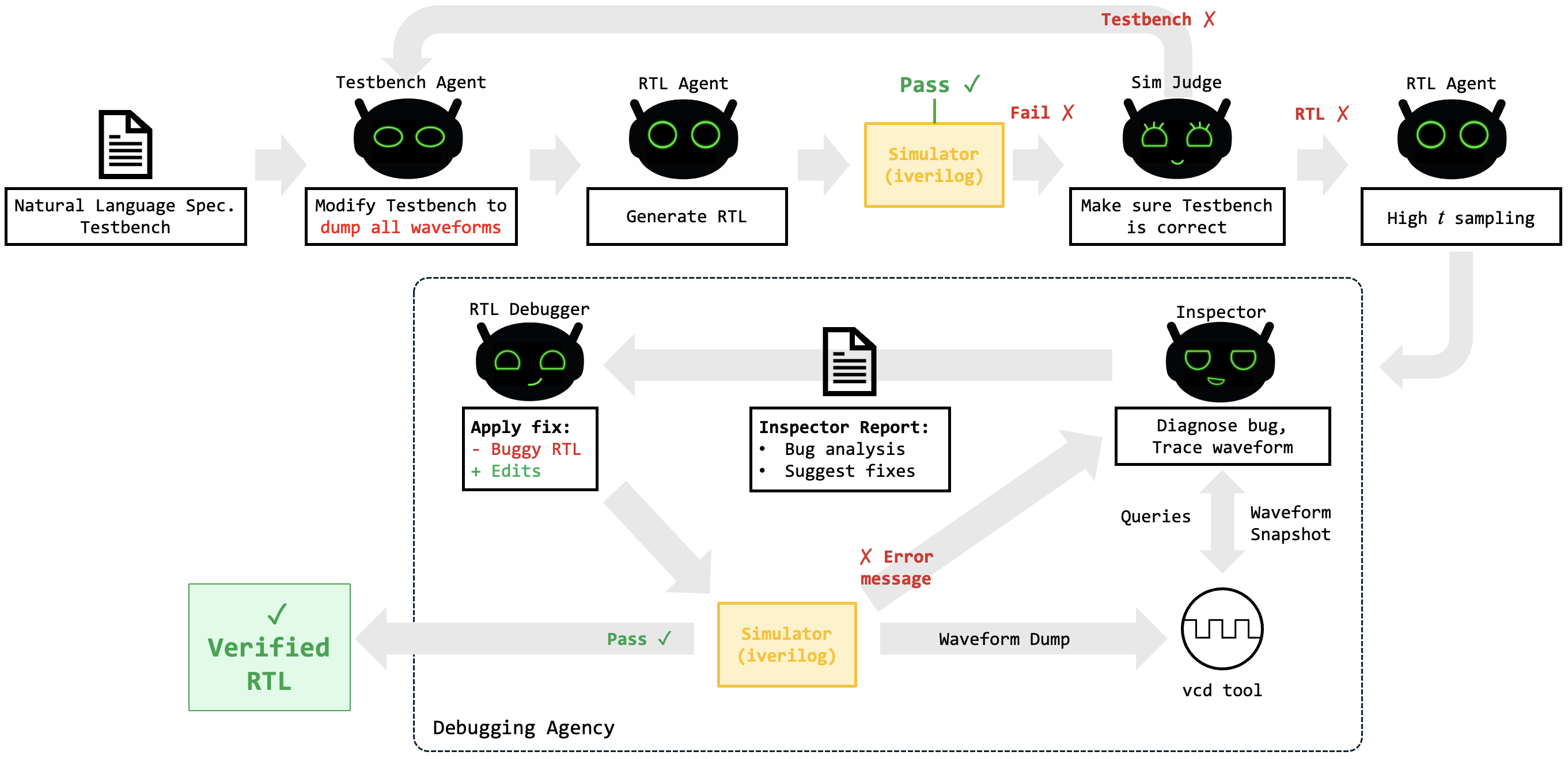}
  \caption{Overview of \name}
  \label{fig: Overview}   
\end{figure*}

\begin{itemize}
    \item We identify \emph{debugging action space completeness} as the key bottleneck in multi-agent RTL generation and introduce Agentic Temporal Exploration, enabling self-directed access to arbitrary signals and time windows for hypothesis-driven code refinement.
    \item We propose \name, a multi-agent system built around Agentic Temporal Exploration with a dedicated Inspector Agent for waveform analysis.
    \item  Rather than ingesting full VCD waveform dumps, the Inspector queries only the signals and time windows relevant to its current hypothesis, reducing token consumption by 18\% and enabling a scalable path toward larger designs.
    \item \name\ achieves \textbf{100\% Pass@1 on VerilogEval-V2}, to the best of our knowledge the first open source system to attain perfect functional correctness on this benchmark.
\end{itemize}

\section{Background}


\subsection{LLMs for RTL Code Generation}
The application of large language models to Verilog generation has progressed along two paths: domain-specialized models trained on curated RTL datasets~\cite{liu2024rtlcoder, zehua2024betterv, deng2025scalertl, chen2021evaluating, liu2023chipnemo, cui2024origen, liu2024craftrtl}, and general-purpose LLMs such as GPT-4 and Claude applied directly to hardware description~\cite{liu2023verilogeval, pinckney2025revisiting}. Both paths demonstrate that LLMs have acquired sufficient understanding of HDL syntax and basic digital design patterns~\cite{thakur2022benchmarkinglargelanguagemodels, pinckney2025revisiting}, but single-pass generation alone, without simulation feedback, remains insufficient for reliable functional correctness\cite{wu2024survey}.

\subsection{Multi-Agent Systems for RTL Generation}
To incorporate simulation feedback, recent work has adopted multi-agent architectures that decompose RTL design into specialized subtasks. VerilogCoder~\cite{ho2025verilogcoder} introduces a Task and Circuit Relation Graph for plan decomposition and an AST-based waveform tracing tool that back-traces signals from mismatched outputs, providing signal-level context at the point of failure. MAGE~\cite{zhao2025mage} employs four agent types with a high-temperature candidate sampling strategy and a state checkpoint mechanism that captures I/O values at the earliest mismatch. ACE-RTL~\cite{deng2026ace} combines a domain-specialized RTL model with a frontier reasoning LLM through an agentic context evolution loop, where a Reflector analyzes simulation logs and a Coordinator maintains structured debugging history across iterations. These systems collectively demonstrate that multi-agent collaboration with simulation-in-the-loop substantially improves functional correctness over single-pass approaches\cite{wu2024autogen, hong2023metagpt, wu2024chateda}.

\subsection{Agentic Reasoning and Tool Use}
\label{{subsec: agent_tool}}
The ReAct paradigm~\cite{yao2022react} formalizes agentic tool use as an interleaved loop of \textit{Thought}, \textit{Action}, and \textit{Observation}: the agent reasons about its current state, invokes a tool, observes the result, and repeats. A key finding from the software engineering domain is that the \emph{interface} between agent and tools matters as much as the underlying LLM\cite{schick2023toolformer}. 

SWE-Agent\cite{yang2024swe} showed that action-space design matters more than model scaling for software agents. Hardware debugging poses an additional challenge: bugs are spatiotemporal, requiring joint control over signal selection and temporal navigation\cite{shinn2023reflexion, wang2024openhands, chen2023teaching, madaan2023self, wang2024executable, wang2023voyager, ul2024aivril}.

\begin{figure*}[t!]
  \centering
  \includegraphics[width=0.95\textwidth]{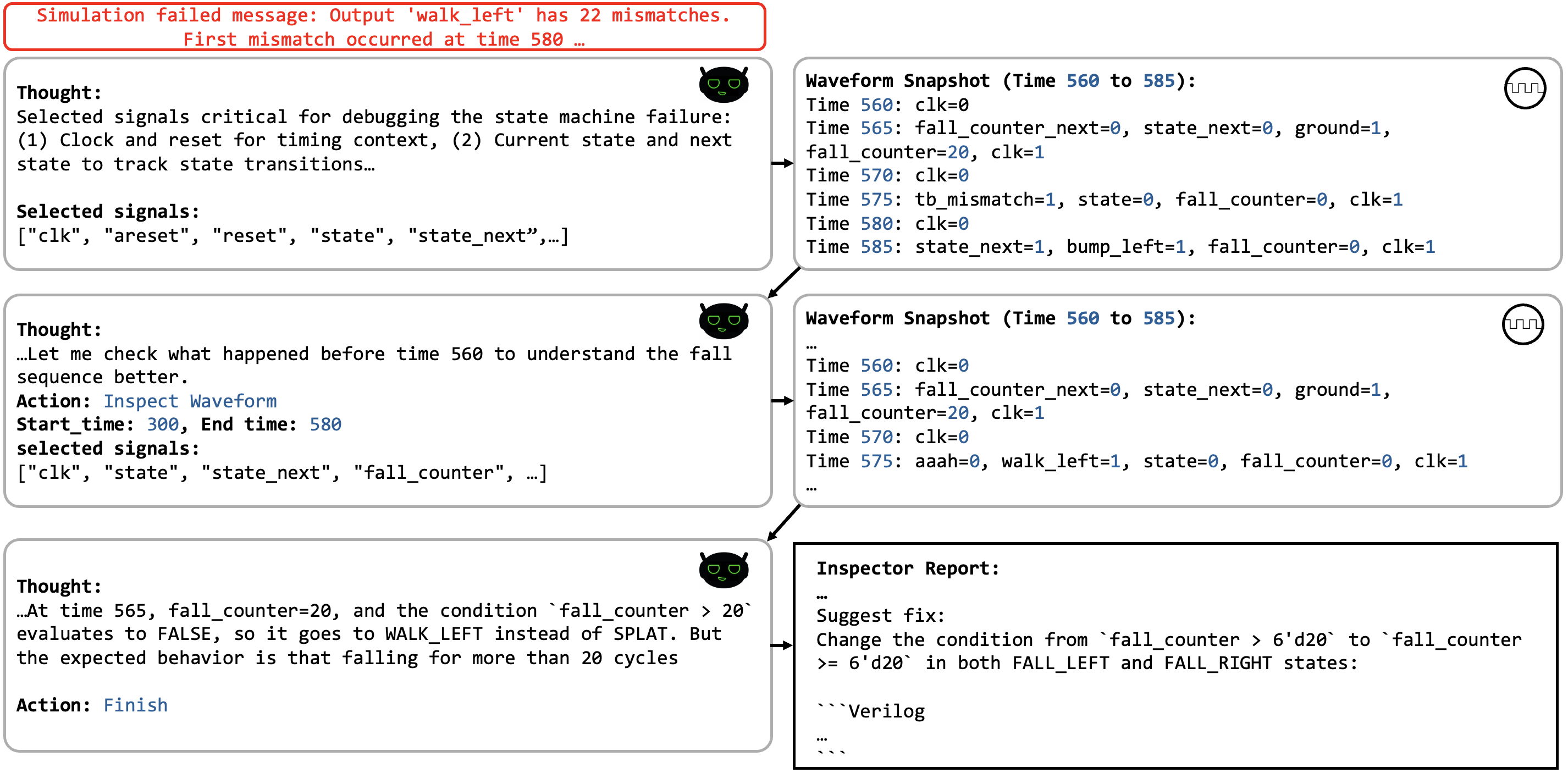}
  \caption{An example of Inspector Agentic Temporal Exploration for Problem 155.}
  \label{fig: Inspector}
\end{figure*}

\section{Method}

VeriTrace's multi-agent architecture follows a deterministic control flow modeled after how human engineers approach RTL design: write code, simulate, inspect failing signals, diagnose the root cause, and repair. Rather than relying on learned orchestration or dynamic agent allocation, VeriTrace decomposes the task into the same sequential stages a verification engineer would follow, with each stage handled by a specialized agent. 

As shown in Figure \ref{fig: Overview}, the system takes a natural language specification and golden testbench if available as input, generates testbench and RTL. Then, if simulation reports a functional error, enters a closed-loop cycle of inspection and debugging until all test cases pass. The key innovation lies in the Inspector agent, which operates over a complete debugging action space to provide hypothesis-driven diagnostic feedback.



\subsection{Testbench and RTL Agent}

The Testbench Agent modifies the golden testbench to dump all waveforms, enabling the Inspector to query arbitrary internal signals, just as human designers do with commercial tools~\cite{synopsys2024interactive}. The RTL Agent generates modules from the specification and modified testbench, with a strict syntax check enforced before simulation. We follow MAGE~\cite{zhao2025mage} to incorporate the Simulation Judge and High-Temperature RTL Sampling for the initial pass, avoiding the token-heavy debugging loop for medium to easy tasks.

\subsection{Inspector: Agentic Temporal Exploration}


Complex RTL problems, particularly FSMs with feedback, are rarely solved in a single generation pass. As established in Section \ref{{subsec: agent_tool}}, RTL bugs are spatiotemporal: a mismatch observed at time $t$ may originate many clock cycles earlier in an unrelated logic block. The Inspector addresses this by separating \textit{diagnosis} from \textit{repair}. While the Debugger translates diagnostic findings into code edits, the Inspector operates over the complete debugging action space, selecting which signals to examine, which time windows to query, and how many rounds of exploration to perform.



This separation of diagnosis from repair\ref{fig: Overview} gives each agent a focused context window, avoiding the need for a single agent to handle waveform analysis and code modification simultaneously.



Figure 2 illustrates how the Inspector, starting from the initial error timestamp, identifies the incorrect state transition through iterative waveform analysis.


\subsubsection{Observation from Human RTL Verification}

When an RTL simulation reports a mismatch, human designers do not examine the failing timestamp in isolation. In commercial tools such as Synopsys Verdi\cite{synopsys2024interactive, verdi}, the designer selects a suspect signal, inspects its transitions over a chosen time window, forms a hypothesis about which logic block introduced the error, and traces further into that block's inputs to confirm or revise the hypothesis. Crucially, it is the designer who decides which signals to examine and when. Diagnosing such faults requires navigating internal signals, state registers, counters, next-state logic, across arbitrary time ranges, iteratively, as understanding evolves.


These observations motivate the Inspector's design. An agent restricted to fixed signals or a single timestamp cannot escape the narrow view that makes automated debugging brittle. The Inspector grants full control over signal selection and time windows, enabling the iterative, hypothesis-driven process formalized below.


\subsubsection{Inspector Action Formulation.}
Let $\mathcal{S}$ denote the natural language specification, $C$ the current RTL code, $t_{\text{err}}$ the first error timestamp reported by simulation, and $\mathcal{W}$ the complete waveform generated by the modified testbench. 

At each iteration $i$, the Inspector selects a set of signals $\boldsymbol{\sigma}_i$ and a time window $[t^{\text{start}}_i,\, t^{\text{end}}_i]$ to query. The agentic waveform tracing tool returns a natural-language-style waveform snapshot:
\begin{equation}
    \omega_i = \textsc{Trace}(\mathcal{W},\, \boldsymbol{\sigma}_i,\, t^{\text{start}}_i,\, t^{\text{end}}_i)
\end{equation}

The Inspector follows a ReAct-style reasoning pattern~\cite{yao2022react}, interleaving thought and action. At each step, the Inspector produces a thought $\tau_i$ that reasons over the accumulated context and decides whether to issue another query or to terminate:

\begin{equation}
    \tau_i,\, a_i = \textsc{Think}(\mathcal{S},\, C,\, t_{\text{err}},\, \omega_1, \dots, \omega_i)
\end{equation}
where $a_i \in \{\textsc{Query},\, \textsc{Finish}\}$. If $a_i = \textsc{Query}$, the Inspector selects new parameters $(\boldsymbol{\sigma}_{i+1},\, t^{\text{start}}_{i+1},\, t^{\text{end}}_{i+1})$ and retrieves the next snapshot $\omega_{i+1}$. This loop continues until either $a_i = \textsc{Finish}$ or a maximum number of iterations $N$ is reached. 

Upon termination, the Inspector produces an \textit{Inspector Report} $\mathcal{R}$ that summarizes the diagnosed behavioral mismatch between $\mathcal{S}$ and $C$, reasoning on the mismatch, and suggested fixes for the subsequent RTL Debugger. Note that the testbench is not exposed to the Inspector since the goal of self-correction should be finding the mismatched behavior between the RTL Code $\mathcal{C}$ and the functional description $\mathcal{S}$ instead of the testbench.

\subsection{RTL Debugger}


After receiving the Inspector Report $\mathcal{R}$, the RTL Debugger analyzes the buggy RTL code $C$ alongside the diagnosed behavioral mismatch and suggested fixes contained in $\mathcal{R}$. The Debugger then performs targeted edits to produce a corrected version $C'$. Rather than attempting to rewrite the entire module, the Debugger focuses on the specific logic blocks identified by the Inspector, preserving the structure of code that is functioning correctly.

The corrected RTL $C'$ is re-simulated. Let $m(C)$ denote the number of output mismatches for a given RTL implementation $C$. If $m(C') < m(C)$, the edit is accepted and $C$ is updated to $C'$. If $m(C') \geq m(C)$, the edit is rolled back to prevent regressions, and the previous version is retained. This checkpoint is necessary because an edit that resolves one mismatch may introduce others, particularly in sequential circuits where shared state logic has cascading effects. Requiring monotonic reduction in mismatch count ensures steady forward progress.


When the accepted edit reduces but does not eliminate all mismatches, the system invokes the Inspector again on the updated code $C'$ with the new error timestamp. This reflects a common pattern in real-world RTL debugging: a single design may contain multiple independent bugs, or fixing one issue may reveal a previously masked failure. The Inspector then begins a fresh round of waveform exploration on the updated design, and the cycle repeats. The full closed-loop iteration between the Inspector and RTL Debugger continues until either $m(C) = 0$ (all test cases pass) or a maximum iteration limit $M$ is reached.





\section{Experiment and Result}


\subsection{Experimental Setup}
We evaluated VeriTrace on VerilogEval-V2 \cite{liu2023verilogeval, pinckney2025revisiting}, a benchmark comprising 156 Verilog coding problems spanning combinational logic and sequential circuits\footnote{One of the problems had a mismatched testbench and prompt. After consulting with the authors of \cite{pinckney2025revisiting}, we made the corresponding adjustments.}. Each problem includes a natural language specification and a golden testbench. We report the highest PASS@1 accuracies from the literature\cite{ho2025verilogcoder, zhao2025mage}. For the reproduced and our method, we report with $n=3$. This mitigates the sampling variance.

\begin{table}[!t]
\centering
\caption{Comparison on VerilogEval-V2.}
\label{tab: main}
\begin{adjustbox}{width=0.95\linewidth}
\begin{tabular}{@{}cccc@{}}
\toprule
\multirow{2}{*}{Method} &
  \multirow{2}{*}{\begin{tabular}[c]{@{}c@{}}Open\\ Souce?\end{tabular}} &
  \multirow{2}{*}{LLM Model} &
  \multirow{2}{*}{\begin{tabular}[c]{@{}c@{}}VerilogEval-V2\\ Pass@1\end{tabular}} \\
             &     &                             &        \\ \midrule
Single Agent & N/A & Claude Sonnet 4.5           & 81.4\% \\
VerilogCoder\cite{ho2025verilogcoder} & \cmark   & GPT4-Turbo                  & 94.2\% \\
ACE-RTL\cite{deng2026ace}      & \xmark   & Custom \& Claude Sonnet 4.0 & 95.5\%$^{3}$ \\
MAGE\cite{zhao2025mage}         & \cmark    & Claude 3.5 Sonnet           & 95.7\% \\
MAGE         & \cmark    & Claude Sonnet 4.0           & 92.3\% \\
Ours         & \cmark    & Claude Sonnet 4.0           & 97.4\% \\
\textbf{Ours}         & \cmark    & Claude Sonnet 4.5           & \textbf{100\%}  \\ \bottomrule
\end{tabular}
\end{adjustbox}
\end{table}

\subsection{Key Results}

Table \ref{tab: main} compares VeriTrace with state-of-the-art open source multi-agent systems \cite{ho2025verilogcoder, zhao2025mage, deng2026ace}. We set N=5 and M=10. With Claude Sonnet 4.5 \cite{anthropic2025sonnet45}, VeriTrace achieves 100\% Pass@1, the first system to achieve perfect functional correctness in VerilogEval-V2, closing the gap that persisted in all prior approaches. To enable a fair same-model comparison, we also evaluate with Claude Sonnet 4.0, where VeriTrace reaches 97.4\% Pass@1, a +5.1\% improvement over MAGE on the identical backbone (92.3\%)\footnote{We reproduced MAGE using its open-source implementation~\cite{zhao2025mage}, substituting Claude Sonnet 4.0 and 4.5 as versions 3.5 and 3.7 have been deprecated. We note significant performance degradation with Claude Sonnet 4.5, attributable to fragile testbench syntax and output parsing.}. This gain is attributable entirely to VeriTrace's debugging agency, as both systems share the same underlying LLM. 

We also note ACE-RTL\cite{deng2026ace}\footnote{ACE-RTL\cite{deng2026ace} reports Agentic Pass Rate (APR), defined as the fraction of uniquely solved problems, rather than the statistical Pass@1 estimator used by other entries. Its implementation is not publicly available.}, which reports 95.5\% APR on VerilogEval-V2 using the same Claude Sonnet 4.0 backbone and its custom fine-tuned model. However, APR measures the fraction of uniquely solved problems across runs rather than the expected single-attempt pass rate, and its closed-source implementation precludes reproduction under controlled conditions. We therefore compare primarily against reproducible open-source baselines.

\begin{table*}
\centering
\caption{Ablation Study}
\label{tab: ab_inspector}
\begin{tabular}{@{}ccccccccc@{}}
\toprule
\multirow{2}{*}{Method} & \multirow{2}{*}{\begin{tabular}[c]{@{}c@{}}VerilogEval-V2\\ Pass@1\end{tabular}} & \multirow{2}{*}{\begin{tabular}[c]{@{}c@{}}Avg. Tokens\\ RTL Agent\end{tabular}} & \multirow{2}{*}{\begin{tabular}[c]{@{}c@{}}Avg. Tokens\\ Testbench Agent\end{tabular}} & \multirow{2}{*}{\begin{tabular}[c]{@{}c@{}}Avg. Tokens\\ Sim Judge\end{tabular}} & \multirow{2}{*}{\begin{tabular}[c]{@{}c@{}}Avg. Tokens\\ RTL Debugger\end{tabular}} & \multirow{2}{*}{\begin{tabular}[c]{@{}c@{}}Avg. Tokens\\ Inspector\end{tabular}} & \multirow{2}{*}{Avg. Tokens} & \multirow{2}{*}{Avg. LLM Call} \\
                        &                                                                                  &                                                                                  &                                                                                        &                                                                                  &                                                                                     &                                                                                  &                              &                                \\ \midrule
w/o Inspector           & 98.29\%                                                                          & 10.2k                                                                            & 6.8k                                                                                   & 1.3k                                                                             & 8.5k                                                                                & N/A                                                                              & 26.7k                        & \textbf{6.4}                            \\
\textbf{w/ Inspector}            & \textbf{100\%}                                                                            & \textbf{10.1k}                                                                            & \textbf{5.4k}                                                                                   & \textbf{0.9k}                                                                             & \textbf{0.4k}                                                                                & \textbf{5.2k}                                                                             & \textbf{21.9k}                        & 6.7                            \\ \bottomrule
\end{tabular}
\end{table*}

\begin{table*}
\centering
\caption{Per-problem ablation on debugging-intensive problems.}
\label{tab: ab_hard}
\begin{tabular}{@{}cccccccccc@{}}
\toprule
\multirow{3}{*}{Problem} & \multirow{3}{*}{Type} & \multicolumn{4}{c}{w/o Inspector} & \multicolumn{4}{c}{w/ Inspector} \\ \cmidrule(l){3-6} \cmidrule(l){7-10}
 &
   &
  \multirow{2}{*}{Pass@1} &
  \multirow{2}{*}{\begin{tabular}[c]{@{}c@{}}Avg. Total\\ Tokens\end{tabular}} &
  \multirow{2}{*}{\begin{tabular}[c]{@{}c@{}}Avg. Debug\\ Tokens\end{tabular}} &
  \multirow{2}{*}{\begin{tabular}[c]{@{}c@{}}Avg. Debug\\ LLM Call\end{tabular}} &
  \multirow{2}{*}{Pass@1} &
  \multirow{2}{*}{\begin{tabular}[c]{@{}c@{}}Avg. Total\\ Tokens\end{tabular}} &
  \multirow{2}{*}{\begin{tabular}[c]{@{}c@{}}Avg. Debug\\ Tokens\end{tabular}} &
  \multirow{2}{*}{\begin{tabular}[c]{@{}c@{}}Avg. Debug\\ LLM Call\end{tabular}} \\
                         &                       &        &         &        &       &        &        &        &       \\ \midrule
Prob062                  & Combinational         & \textbf{100\%}  & \textbf{41.7k}   & \textbf{4k}     & \textbf{2.00}  & \textbf{100\%}   & 44.6k  & 7.3k   & 4.33  \\
Prob070                  & Combinational         & \textbf{100\%}   & 64.7k   & 10.5k  & \textbf{2.00}  & \textbf{100\%}   & \textbf{63.2k}  & \textbf{6.3k}   & 4.00  \\
Prob078                  & Sequential            & \textbf{100\%}   & 50.6k   & 18.4k  & \textbf{4.67}  & \textbf{100\%}   & \textbf{46.8k}  & \textbf{7.6k}   & 5.00  \\
Prob093                  & Combinational         & \textbf{100\%}   & 94.7k   & 47.2k  & \textbf{4.67}  & \textbf{100\%}   & \textbf{83.1k}  & \textbf{36.5k}  & 8.33  \\
Prob137                  & FSM (Seq.)            & 0\%    & \textbf{348.2k}  & \textbf{301.7k} & \textbf{30.00} & \textbf{100\%}   & 383.6k & 339.6k & 56.00 \\
Prob149                  & FSM (Seq.)            & 33\%   & 345.9k  & 281.2k & \textbf{20.67} & \textbf{100\%}   & \textbf{317.4k} & \textbf{256.9k} & 31.67 \\
Prob155                  & FSM (Seq.)            & 33\%   & 619.5k  & 530.8k & 27.33 & \textbf{100\%}  & \textbf{110.3k} & \textbf{32.6k}  & \textbf{9.00}  \\ \bottomrule
\end{tabular}
\end{table*}

\subsection{Ablation Studies}

\subsubsection{The Effectiveness of Inspector}

To isolate the contribution of the Inspector, we compare VeriTrace with and without this component. Table \ref{tab: ab_inspector} reports per-agent and total token consumption, averaged across all problems. Because LLM API latency introduces variability in wall-clock time, we report the average number of LLM calls as a proxy for runtime instead. Pass@1 is computed over $n=3$ runs to reduce sampling variance. Both configurations are using the Claude Sonnet 4.5 backbone\cite{anthropic2025sonnet45} with temperature $t=0.85$.

Both configurations consume comparable tokens during RTL generation, confirming that the Inspector's benefit arises in the debugging phase rather than initial code generation. With Inspector, waveform analysis is offloaded to a dedicated agent that produces targeted diagnostic reports, enabling the RTL Debugger to apply more precise fixes. This specialization reduces combined debugging tokens (RTL Debugger + Inspector) relative to the RTL Debugger alone without the Inspector, yielding an overall reduction of approximately 18\% in total token consumption. This improvement comes with a trade-off: due to the nature of iterative inspection, \name \ uses more rounds of LLM calls on average, leading to longer runtime.

\subsubsection{Analysis on the Hard Problems}
To understand which problem categories benefit the most, we further examine the 7 problems that consistently trigger the debugging loop in both configurations (Table \ref{tab: ab_hard}). For combinational and non-FSM sequential circuits, both systems achieve 100\% Pass@1. The Inspector's advantage here is primarily efficiency as it reduces debugging tokens in 3 of 4 cases by providing more actionable diagnostic information upfront. 

The contrast is sharpest on FSM problems, where the system without the Inspector achieves only 0–33\% Pass@1. The Inspector's ability to trace state transitions across time windows allows it to pinpoint root causes in feedback-heavy designs, boosting all three to 100\% Pass@1. Notably, for Prob155 the Inspector reduces total token consumption by over 5× (619k → 110k), demonstrating that accurate diagnosis not only improves correctness but also avoids the wasteful trial-and-error cycles of blind retry. In this case we observed that the RTL Debugger ended up editing the whole module after multiple failed attempts, consuming significantly more tokens.

\section{Discussion and Limitations}


VerilogEval-V2 comprises 156 single-module problems with provided golden testbenches. Industrial designs involve multi-module hierarchies, incomplete specifications, and far larger signal spaces. Whether the Inspector's temporal exploration scales to such settings remains an open question. Multi-task benchmarks such as CVDP~\cite{pinckney2025comprehensiveverilogdesignproblems, deng2026ace, allam2024rtl} would exercise the framework under more diverse conditions. Because CVDP evaluation requires multiple independent runs per problem with frontier-model API calls at each iteration, the associated cost is non-trivial for academic settings. In preliminary experiments, we applied Agentic Temporal Exploration within a hierarchical multi-agent framework to two challenging CVDP specification-to-RTL problems and solved both, suggesting that the debugging methodology may generalize beyond single-module benchmarks. Comprehensive evaluation is planned as follow-up work.



The Testbench Agent dumps waveforms for all signals by default, which is what enables the Inspector to query arbitrary internal signals. Critically, the Inspector does not ingest the full VCD into the LLM context. Instead, it retrieves only the signals and time windows it selects at each iteration, so token cost scales with diagnostic complexity rather than design size. For the single-module problems in VerilogEval-V2 this storage overhead is negligible, but for industrial-scale modules with thousands of signals, adaptive waveform dumping or on-demand re-simulation may be necessary. 

Finally, as evidenced by the gap between Claude Sonnet 4.0 (97.4\%) and 4.5 (100\%), all current multi-agent RTL systems remain sensitive to the underlying LLM. Disentangling architectural from model-level contributions remains a shared challenge.

\section{Conclusion}

We presented \name, a multi-agent system that introduces Agentic Temporal Exploration, granting LLM agents full control over signal selection and temporal exploration during debugging. VeriTrace achieves 100\% Pass@1 on VerilogEval-V2, the first perfect score on this benchmark, while reducing token consumption by ~18\% through more targeted diagnosis. Our results demonstrate that debugging agency, not just richer feedback, is the key to closing the final accuracy gap in LLM-based RTL generation. Future work will extend Agentic Temporal Exploration to multi-module designs and dynamic agent allocation. The implementation of VeriTrace will be publicly available.

\pagebreak

\bibliographystyle{IEEEbib}
\bibliography{references}

@inproceedings{ho2025verilogcoder,
  title={Verilogcoder: Autonomous verilog coding agents with graph-based planning and abstract syntax tree (ast)-based waveform tracing tool},
  author={Ho, Chia-Tung and Ren, Haoxing and Khailany, Brucek},
  booktitle={Proceedings of the AAAI Conference on Artificial Intelligence},
  volume={39},
  number={1},
  pages={300--307},
  year={2025}
}

@inproceedings{liu2023verilogeval,
  title={Verilogeval: Evaluating large language models for verilog code generation},
  author={Liu, Mingjie and Pinckney, Nathaniel and Khailany, Brucek and Ren, Haoxing},
  booktitle={2023 IEEE/ACM International Conference on Computer Aided Design (ICCAD)},
  pages={1--8},
  year={2023},
  organization={IEEE}
}

@article{pinckney2025revisiting,
  title={Revisiting verilogeval: A year of improvements in large-language models for hardware code generation},
  author={Pinckney, Nathaniel and Batten, Christopher and Liu, Mingjie and Ren, Haoxing and Khailany, Brucek},
  journal={ACM Transactions on Design Automation of Electronic Systems},
  year={2025},
  publisher={ACM New York, NY}
}

@inproceedings{zhao2025mage,
  title={Mage: A multi-agent engine for automated rtl code generation},
  author={Zhao, Yujie and Zhang, Hejia and Huang, Hanxian and Yu, Zhongming and Zhao, Jishen},
  booktitle={2025 62nd ACM/IEEE Design Automation Conference (DAC)},
  pages={1--7},
  year={2025},
  organization={IEEE}
}

@article{yao2022react,
  title={ReAct: Synergizing Reasoning and Acting in Language Models},
  author={Yao, Shunyu and Zhao, Jeffrey and Yu, Dian and Du, Nan and Shafran, Izhak and Narasimhan, Karthik and Cao, Yuan},
  journal={arXiv preprint arXiv:2210.03629},
  year={2022}
}

@misc{verdi,
  author = {{Synopsys, Inc.}},
  title = {Verdi Automated Debug System},
  howpublished = {\url{https://www.synopsys.com/verification/debug/verdi.html}},
  year = {2024},
  note = {Accessed: 2025-02-05}
}

@INPROCEEDINGS{choi2013hls,
  author={Choi, Jongsok and Brown, Stephen and Anderson, Jason},
  booktitle={2013 International Conference on Field-Programmable Technology (FPT)}, 
  title={From software threads to parallel hardware in high-level synthesis for FPGAs}, 
  year={2013},
  volume={},
  number={},
  pages={270-277},
  doi={10.1109/FPT.2013.6718365}}

@INPROCEEDINGS{faber2022challenges,
  author={Faber, Clayton J. and Harris, Steven D. and Xiac, Zhili and Chamberlain, Roger D. and Cabrera, Anthony M.},
  booktitle={2022 IEEE High Performance Extreme Computing Conference (HPEC)}, 
  title={Challenges Designing for FPGAs Using High-Level Synthesis}, 
  year={2022},
  volume={},
  number={},
  pages={1-7},
  doi={10.1109/HPEC55821.2022.9926398}}

@article{pan2025survey,
  title={A survey of research in large language models for electronic design automation},
  author={Pan, Jingyu and Zhou, Guanglei and Chang, Chen-Chia and Jacobson, Isaac and Hu, Jiang and Chen, Yiran},
  journal={ACM Transactions on Design Automation of Electronic Systems},
  volume={30},
  number={3},
  pages={1--21},
  year={2025},
  publisher={ACM New York, NY}
}

@techreport{anthropic2025sonnet45,
  author = {{Anthropic}},
  title = {System Card: Claude Sonnet 4.5},
  year = {2025},
  month = {September},
  institution = {Anthropic},
  url = {https://www.anthropic.com/research/claude-sonnet-4-5-system-card}
}

@misc{thakur2022benchmarkinglargelanguagemodels,
      title={Benchmarking Large Language Models for Automated Verilog RTL Code Generation}, 
      author={Shailja Thakur and Baleegh Ahmad and Zhenxing Fan and Hammond Pearce and Benjamin Tan and Ramesh Karri and Brendan Dolan-Gavitt and Siddharth Garg},
      year={2022},
      eprint={2212.11140},
      archivePrefix={arXiv},
      primaryClass={cs.PL},
      url={https://arxiv.org/abs/2212.11140}, 
}

@misc{tsai2024rtlfixerautomaticallyfixingrtl,
      title={RTLFixer: Automatically Fixing RTL Syntax Errors with Large Language Models}, 
      author={Yun-Da Tsai and Mingjie Liu and Haoxing Ren},
      year={2024},
      eprint={2311.16543},
      archivePrefix={arXiv},
      primaryClass={cs.AR},
      url={https://arxiv.org/abs/2311.16543}, 
}

@inproceedings{lu2024rtllm,
  author={Lu, Yao and Liu, Shang and Zhang, Qijun and Xie, Zhiyao},
  booktitle={2024 29th Asia and South Pacific Design Automation Conference (ASP-DAC)}, 
  title={RTLLM: An Open-Source Benchmark for Design RTL Generation with Large Language Model}, 
  year={2024},
  pages={722-727},
  organization={IEEE}
  }

@online{synopsys2024interactive,
  author={{Synopsys}},
  title={Verdi Debug: Streamlining Verification Engineers' Workflow},
  year={2024},
  url={https://www.synopsys.com/blogs/chip-design/interactive-debugging-reducing-simulation-debug-tat.html},
  urldate={2026-03-04},
  note={Synopsys Blog}
}

@article{liu2024rtlcoder,
  title={Rtlcoder: Fully open-source and efficient llm-assisted rtl code generation technique},
  author={Liu, Shang and Fang, Wenji and Lu, Yao and Wang, Jing and Zhang, Qijun and Zhang, Hongce and Xie, Zhiyao},
  journal={IEEE Transactions on Computer-Aided Design of Integrated Circuits and Systems},
  volume={44},
  number={4},
  pages={1448--1461},
  year={2024},
  publisher={IEEE}
}

@inproceedings{zehua2024betterv,
  title={Betterv: Controlled verilog generation with discriminative guidance},
  author={Zehua, PEI and Zhen, Huiling and Yuan, Mingxuan and Huang, Yu and Yu, Bei},
  booktitle={Forty-first International Conference on Machine Learning},
  year={2024}
}

@inproceedings{deng2025scalertl,
  title={Scalertl: Scaling llms with reasoning data and test-time compute for accurate rtl code generation},
  author={Deng, Chenhui and Tsai, Yun-Da and Liu, Guan-Ting and Yu, Zhongzhi and Ren, Haoxing},
  booktitle={2025 ACM/IEEE 7th Symposium on Machine Learning for CAD (MLCAD)},
  pages={1--9},
  year={2025},
  organization={IEEE}
}

@article{deng2026ace,
  title={ACE-RTL: When Agentic Context Evolution Meets RTL-Specialized LLMs},
  author={Deng, Chenhui and Yu, Zhongzhi and Liu, Guan-Ting and Pinckney, Nathaniel and Ren, Haoxing},
  journal={arXiv preprint arXiv:2602.10218},
  year={2026}
}

@article{yang2024swe,
  title={Swe-agent: Agent-computer interfaces enable automated software engineering},
  author={Yang, John and Jimenez, Carlos E and Wettig, Alexander and Lieret, Kilian and Yao, Shunyu and Narasimhan, Karthik and Press, Ofir},
  journal={Advances in Neural Information Processing Systems},
  volume={37},
  pages={50528--50652},
  year={2024}
}

@misc{pinckney2025comprehensiveverilogdesignproblems,
      title={Comprehensive Verilog Design Problems: A Next-Generation Benchmark Dataset for Evaluating Large Language Models and Agents on RTL Design and Verification}, 
      author={Nathaniel Pinckney and Chenhui Deng and Chia-Tung Ho and Yun-Da Tsai and Mingjie Liu and Wenfei Zhou and Brucek Khailany and Haoxing Ren},
      year={2025},
      eprint={2506.14074},
      archivePrefix={arXiv},
      primaryClass={cs.LG},
      url={https://arxiv.org/abs/2506.14074}, 
}

@article{schick2023toolformer,
  title={Toolformer: Language models can teach themselves to use tools},
  author={Schick, Timo and Dwivedi-Yu, Jane and Dess{\`\i}, Roberto and Raileanu, Roberta and Lomeli, Maria and Hambro, Eric and Zettlemoyer, Luke and Cancedda, Nicola and Scialom, Thomas},
  journal={Advances in neural information processing systems},
  volume={36},
  pages={68539--68551},
  year={2023}
}

@article{shinn2023reflexion,
  title={Reflexion: Language agents with verbal reinforcement learning},
  author={Shinn, Noah and Cassano, Federico and Gopinath, Ashwin and Narasimhan, Karthik and Yao, Shunyu},
  journal={Advances in neural information processing systems},
  volume={36},
  pages={8634--8652},
  year={2023}
}

@inproceedings{wu2024autogen,
  title={Autogen: Enabling next-gen LLM applications via multi-agent conversations},
  author={Wu, Qingyun and Bansal, Gagan and Zhang, Jieyu and Wu, Yiran and Li, Beibin and Zhu, Erkang and Jiang, Li and Zhang, Xiaoyun and Zhang, Shaokun and Liu, Jiale and others},
  booktitle={First conference on language modeling},
  year={2024}
}

@inproceedings{hong2023metagpt,
  title={MetaGPT: Meta programming for a multi-agent collaborative framework},
  author={Hong, Sirui and Zhuge, Mingchen and Chen, Jonathan and Zheng, Xiawu and Cheng, Yuheng and Wang, Jinlin and Zhang, Ceyao and Wang, Zili and Yau, Steven Ka Shing and Lin, Zijuan and others},
  booktitle={The twelfth international conference on learning representations},
  year={2023}
}

@article{wang2024openhands,
  title={Openhands: An open platform for ai software developers as generalist agents},
  author={Wang, Xingyao and Li, Boxuan and Song, Yufan and Xu, Frank F and Tang, Xiangru and Zhuge, Mingchen and Pan, Jiayi and Song, Yueqi and Li, Bowen and Singh, Jaskirat and others},
  journal={arXiv preprint arXiv:2407.16741},
  year={2024}
}

@inproceedings{wang2024executable,
  title={Executable code actions elicit better llm agents},
  author={Wang, Xingyao and Chen, Yangyi and Yuan, Lifan and Zhang, Yizhe and Li, Yunzhu and Peng, Hao and Ji, Heng},
  booktitle={Forty-first International Conference on Machine Learning},
  year={2024}
}

@article{wang2023voyager,
  title={Voyager: An open-ended embodied agent with large language models},
  author={Wang, Guanzhi and Xie, Yuqi and Jiang, Yunfan and Mandlekar, Ajay and Xiao, Chaowei and Zhu, Yuke and Fan, Linxi and Anandkumar, Anima},
  journal={arXiv preprint arXiv:2305.16291},
  year={2023}
}

@article{chen2021evaluating,
  title={Evaluating large language models trained on code},
  author={Chen, Mark and Tworek, Jerry and Jun, Heewoo and Yuan, Qiming and Pinto, Henrique Ponde De Oliveira and Kaplan, Jared and Edwards, Harri and Burda, Yuri and Joseph, Nicholas and Brockman, Greg and others},
  journal={arXiv preprint arXiv:2107.03374},
  year={2021}
}

@article{chen2023teaching,
  title={Teaching large language models to self-debug},
  author={Chen, Xinyun and Lin, Maxwell and Sch{\"a}rli, Nathanael and Zhou, Denny},
  journal={arXiv preprint arXiv:2304.05128},
  year={2023}
}

@article{liu2023chipnemo,
  title={Chipnemo: Domain-adapted llms for chip design},
  author={Liu, Mingjie and Ene, Teodor-Dumitru and Kirby, Robert and Cheng, Chris and Pinckney, Nathaniel and Liang, Rongjian and Alben, Jonah and Anand, Himyanshu and Banerjee, Sanmitra and Bayraktaroglu, Ismet and others},
  journal={arXiv preprint arXiv:2311.00176},
  year={2023}
}

@inproceedings{cui2024origen,
  title={Origen: Enhancing rtl code generation with code-to-code augmentation and self-reflection},
  author={Cui, Fan and Yin, Chenyang and Zhou, Kexing and Xiao, Youwei and Sun, Guangyu and Xu, Qiang and Guo, Qipeng and Liang, Yun and Zhang, Xingcheng and Song, Demin and others},
  booktitle={Proceedings of the 43rd IEEE/ACM International Conference on Computer-Aided Design},
  pages={1--9},
  year={2024}
}

@article{liu2024craftrtl,
  title={Craftrtl: High-quality synthetic data generation for verilog code models with correct-by-construction non-textual representations and targeted code repair},
  author={Liu, Mingjie and Tsai, Yun-Da and Zhou, Wenfei and Ren, Haoxing},
  journal={arXiv preprint arXiv:2409.12993},
  year={2024}
}

@article{wu2024survey,
  title={Survey of machine learning for software-assisted hardware design verification: Past, present, and prospect},
  author={Wu, Nan and Li, Yingjie and Yang, Hang and Chen, Hanqiu and Dai, Steve and Hao, Cong and Yu, Cunxi and Xie, Yuan},
  journal={ACM Transactions on Design Automation of Electronic Systems},
  volume={29},
  number={4},
  pages={1--42},
  year={2024},
  publisher={ACM New York, NY}
}

@article{wu2024chateda,
  title={Chateda: A large language model powered autonomous agent for eda},
  author={Wu, Haoyuan and He, Zhuolun and Zhang, Xinyun and Yao, Xufeng and Zheng, Su and Zheng, Haisheng and Yu, Bei},
  journal={IEEE Transactions on Computer-Aided Design of Integrated Circuits and Systems},
  volume={43},
  number={10},
  pages={3184--3197},
  year={2024},
  publisher={IEEE}
}

@article{ul2024aivril,
  title={Aivril: Ai-driven rtl generation with verification in-the-loop},
  author={ul Islam, Mubashir and Sami, Humza and Gaillardon, Pierre-Emmanuel and Tenace, Valerio and others},
  journal={arXiv preprint arXiv:2409.11411},
  year={2024}
}

@article{madaan2023self,
  title={Self-refine: Iterative refinement with self-feedback},
  author={Madaan, Aman and Tandon, Niket and Gupta, Prakhar and Hallinan, Skyler and Gao, Luyu and Wiegreffe, Sarah and Alon, Uri and Dziri, Nouha and Prabhumoye, Shrimai and Yang, Yiming and others},
  journal={Advances in neural information processing systems},
  volume={36},
  pages={46534--46594},
  year={2023}
}

@inproceedings{allam2024rtl,
  title={Rtl-repo: A benchmark for evaluating llms on large-scale rtl design projects},
  author={Allam, Ahmed and Shalan, Mohamed},
  booktitle={2024 IEEE LLM Aided Design Workshop (LAD)},
  pages={1--5},
  year={2024},
  organization={IEEE}
}
\end{document}